\documentclass[final,3p,times,twocolumn]{elsarticle}

\journal{}
    
\usepackage{amsmath}
\usepackage{xcolor}
\usepackage{color} 
\usepackage{multirow}
\usepackage{epstopdf}
\usepackage{graphicx}
\usepackage{url}
\usepackage{adjustbox}

\begin{document}

\begin{frontmatter}

\title{Optimal Gait Control for a Tendon-driven Soft Quadruped Robot by Model-based Reinforcement Learning}

\author[1address]{Xuezhi Niu\corref{contrib}}
\author[1address]{Kaige Tan\corref{contrib}}
\author[1address]{Lei Feng\corref{mycorrespondingauthor}}
\cortext[mycorrespondingauthor]{Corresponding author}
\cortext[contrib]{Authors contributed equally}
\ead{lfeng@kth.se}

\address[1address]{Department of Engineering Design, KTH Royal Institute of Technology, Stockholm 10044, Sweden}

\begin{abstract}
This study presents an innovative approach to optimal gait control for a soft quadruped robot enabled by four Compressible Tendon-driven Soft Actuators (CTSAs). Improving our previous studies of using model-free reinforcement learning for gait control, we employ model-based reinforcement learning (MBRL) to further enhance the performance of the gait controller. Compared to rigid robots, the proposed soft quadruped robot has better safety, less weight, and a simpler mechanism for fabrication and control. However, the primary challenge lies in developing sophisticated control algorithms to attain optimal gait control for fast and stable locomotion. The research employs a multi-stage methodology, including state space restriction, data-driven model training, and reinforcement learning algorithm development. Compared to benchmark methods, the proposed MBRL algorithm, combined with post-training, significantly improves the efficiency and performance of gait control policies. The developed policy is both robust and adaptable to the robot's deformable morphology. The study concludes by highlighting the practical applicability of these findings in real-world scenarios.
\end{abstract}


\begin{keyword}
Quadruped robot\sep Soft actuators\sep Reinforcement learning\sep Motion control
\end{keyword}

\end{frontmatter}

\section{Introduction}

Legged robots have demonstrated remarkable agility in both academia~\cite{choi2023learning, bledt2018MIT, hutter2016anymal} and industry\footnote{Unitree: \url{https://www.unitree.com/go2}}$^,$\footnote{Boston Dynamics: \url{https://bostondynamics.com/products/spot/}}. Most of them were made of rigid materials by convention, but the rigid robots suffered from low energy density or require heavy onboard power sources for extended operation~\cite{taheri2023Studya}. In contrast, soft robots offer distinct advantages, including lightweight, compliance, and affordability~\cite{yasa2023Overview}. These features have made soft robots ideal for collaborative tasks in healthcare, search and rescue, human-robot interaction scenarios, and omnidirectional navigation~\cite{drotman20173d,ji2022omnidirectional}. Owing to the nonlinear dynamics and infinite degree of freedom (DoF) of soft materials, traditional control methods based on dynamic models of the robots become inadequate. This underscores the need for advanced control strategies tailored to the ever-changing morphology of soft robots~\cite{wang2022control}, differentiating them from traditional rigid robots~\cite{taheri2023Studya}. In particular, rigid quadruped robots, a common type of legged robot mimicking the locomotion of four-legged animals, are proficient in agile locomotion~\cite{taheri2023Studya}. Soft quadruped robots have the advantage of safe interaction with the environment thanks to flexible and deformable structures~\cite{rus2015design}.

Historically, locomotion control for rigid-legged robots usually relied on model-based optimal control approaches such as Whole-Body Control (WBC)~\cite{fahmi2019passive, dudzik2020robust}, Model Predictive Control (MPC)~\cite{sleiman2021unified, dicarlo2018Dynamic}, and Trajectory Optimization (TO)~\cite{ponton2021efficient}. Recently, there has been growing interest in leveraging Reinforcement Learning (RL) techniques to enhance the control and adaptability of soft robotic systems through iterative improvements~\cite{miki2022learning, tsounis2020deepgait}. The selection for the best control design method is still ad hoc. As the complexity of robotics increases, RL is increasingly important to develop robust control algorithms for robots with many DoFs and time-variant properties. In RL, learning algorithms, namely agents, optimize control policies based on performance feedback measured by the robots, which are often called environments. The environment may be either a physical robot or more commonly a simulation model of the physical robot. 
To precisely capture the detailed dynamics of the physical robot, the simulation model is often complex and time-consuming for execution. RL algorithms that directly obtain samples and rewards from the complex simulation models are also inefficient and are often called \emph{model-free RL} (MFRL) methods~\cite{morimoto2021ModelFree, ji2022synthesizing}. 

To improve the training efficiency of RL, researchers have explored various strategies that improve the traditional MFRL methods. One prominent approach is the use of \emph{model-based RL} (MBRL), where a simplified and computationally efficient approximate model replaces the original environment model. MBRL can significantly reduce the sample complexity compared to MFRL, thus enhancing efficiency and stability in the learning process~\cite{thuruthel2019ModelBased, pei2021improved, yu2023safe}. In addition, \emph{Meta RL} is also promising to boost RL efficiency~\cite{bing2023meta, ballou2023variational}. This approach focuses on training models to learn how to learn, enabling them to quickly adapt to new tasks with minimal data. It often operates by training on the distribution of tasks and optimizing for generalization across them. Moreover, \emph{exploration strategies} are crucial in enhancing the efficiency of RL. Methods such as intrinsic motivation~\cite{10006753}, curiosity-driven exploration~\cite{li2020random}, and uncertainty-based exploration~\cite{jiang2024importance} enable the agent to identify informative and rewarding states more effectively than relying on simple random exploration.

Specifically, MBRL is gaining interest in the robotics field, mainly due to data efficiency challenges and physical hardware constraints~\cite{ibarz2021How}. The approximated model can be used to simulate future states and rewards, thus allowing the agent to plan and make decisions based on anticipated outcomes without relying on direct interaction with the environment. Therefore, substantial data samples and rewards can be obtained quickly from the approximation model, and the training process can be significantly expedited. Several neural network approaches are applicable for the approximation model, including deep neural network (DNN)~\cite{xie2023data}, recurrent neural network (RNN)~\cite{cao2021learning}, and convolutional neural network (CNN)~\cite{claessens2016convolutional}. Since CNNs are effective for identifying spatial patterns and features, they are suitable for spatial analysis instead of complex temporal relationship, and hence not applicable for modeling dynamics. RNNs, on the other hand, incorporate loops in the network structure to store information from previous inputs and use it to generate the next output in the sequence. They are hence effective for handling sequences, time series data, and dynamic patterns
~\cite{lipton2015critical, sherstinsky2020fundamentals}. DNNs can also model complex relationships through multiple layers of abstraction. They are also applicable for modeling the dynamics of high-dimensional systems, such as robot locomotion dynamics.

In our previous studies~\cite{ji2022omnidirectional, ji2022synthesizing}, we developed SoftQ (\underline{Soft} \underline{Q}uadruped), a soft quadruped robot enabled by four Compressible Tendon-driven Soft Actuators (CTSAs) as legs, each controlled by three servo motors. To avoid learning directly with the physical robot, a dynamics model of the SoftQ has been made by SimScape Multibody. The Soft Actor-Critic (SAC) algorithm is employed to learn an optimal gait of the SoftQ. However, two limitations are observed. First, the SimScape model of the SoftQ is accurate but time-consuming, necessitating around 30 seconds of computation for each elapsed physical second. Second, an MFRL approach is applied in the prior work, where the end-to-end control is performed by directly using the motor output as the action space. Despite its popularity in RL applications, the MFRL method is inefficient for mobile robots due to the generation of numerous unfeasible or redundant leg motor position combinations, by which the learning process does not exploit the model knowledge of the robot.

This study investigates gait control for SoftQ through a customized MBRL approach by incorporating \emph{post-training} (PT). In MBRL, a \emph{surrogate model} of the environment is learned to approximate the dynamics of the environment with much less computation time~\cite{ibarz2021How}. The learning agent employs this surrogate model to estimate the next state and corresponding reward for a given current state and action, mitigating the computational burden of direct interaction with the environment. Despite the computational efficiency of MBRL, the inherent divergence between the real and surrogate models may lead to sub-optimality in the converged controller. Consequently, PT is applied subsequently to enhance the RL controller's performance. PT shares similarities with \emph{continual learning}~\cite{huang2021continual}, which takes place after the convergence of an initial training process using the surrogate model and exposes the converged controller to the real environment. The method enables an additional training phase to refine its capabilities for real-world adaptation. Subsequently, PT facilitates the transfer of acquired skills from MBRL, allowing them to be adapted and applied to the real robot.

In summary, the contributions of this article are as follows: (i) We investigate the MBRL approach with a data-driven surrogate model for training gait controllers in soft quadruped robots. Our approach guarantees the state transition estimation accuracy while improving the training efficiency compared to the traditional model-free method. (ii) By employing a parametric gait pattern model, we effectively reduce the state and action spaces during the exploration in MBRL, thereby enhancing the control system's efficiency for the SoftQ robot. (iii) We provide empirical evidence of the real-world effectiveness of our proposed algorithms, highlighting the potential applicability and adaptability of our methodology for optimizing gait control.

The rest of this article is organized as follows. Section~\ref{sec:preliminary} shows the background and an overview of the reinforcement learning framework. Section~\ref{sec:data_driven} introduces the studied soft quadruped robot and develops its surrogate model using a data-driven approach. Section~\ref{sec:MBRL} elaborates on the MBRL algorithm and emphasizes its real-world performance improvement for robotic systems. Section~\ref{sec:result_validation} outlines the hardware implementation, experiment setup, and evaluation results. Finally, Section~\ref{sec:conclusion} provides the conclusion for this article.

\section{Preliminaries}
\label{sec:preliminary}
The reinforcement learning framework is defined by an agent interacting with its environment at the discrete time steps $t$. The agent observes a state $\mathbf{s}_t\in\mathcal{S}$, performs an action $\mathbf{a}_t\in\mathcal{A}$, receives a reward $r_t\in\mathbb{R}$, and transitions to a new state $\mathbf{s}_{t+1}$, as dictated by the environment's dynamics $f: \mathcal{S} \times \mathcal{A} \rightarrow \mathcal{S}$. In MBRL, we approximate this dynamics function with a surrogate model $\hat{f}_\theta$, parameterized by $\theta$, to predict state transitions. We assume $\hat{f}_\theta$ follows a Markov Decision Process ($\mathcal{S}, \mathcal{A}, P, r$) with continuous state ($\mathcal{S}$) and action ($\mathcal{A}$) spaces. The transition function $P(\mathbf{s}_{t+1}\!\mid\!\mathbf{s}_t, \mathbf{a}_t)$ evaluates the probability density of transitioning from the current state $\mathbf{s}_t$ to the subsequent state $\mathbf{s}_{t+1}$ upon executing action $\mathbf{a}_t$. After taking action $\mathbf{a}_t$ in state $\mathbf{s}_t$, the agent immediately receives a reward $r_t := r(\mathbf{s}_{t}, \mathbf{a}_{t})$ from the reward function $r: \mathcal{S} \times \mathcal{A} \to \mathbb{R}$, quantifying action desirability.

\subsection{Soft Actor-Critic}
SAC algorithm uses the surrogate model for the optimal gait policy training, which is selected because of its superior performance in continuous action space environments~\cite{haarnoja2018soft}. It is designed to derive a policy that maximizes cumulative expected rewards while considering policy distribution entropy~\cite{haarnoja2019Soft}. Policy entropy, denoted by $H(\pi(\cdot\!\mid\!\mathbf{s}_t)) = \mathbb{E}_{\mathbf{a}_t\sim \pi}[-\ln \pi(\mathbf{a}_t\!\mid\!\mathbf{s}_t)]$, quantifies the uncertainty of the action selection for a policy $\pi$ in state $\mathbf{s}_t$. Higher policy entropy promotes exploration during training, thereby mitigating the risk of local policy convergence. The training is terminated upon reaching a predetermined reward threshold or the training episode reaches a predefined limit.

The SAC algorithm aims to find a stochastic policy $\pi$ that maximizes a trade-off between expected return and entropy. The value function $V_\pi(\mathbf{s}_t)$ of a state by following the policy $\pi$ is regarded as the entropy-constrained expectation of the accumulated reward, and
\begin{equation}
\label{eq:optimalSAC}
\begin{aligned}
    &\pi^* = \arg\max_\pi V_\pi(\mathbf{s}_t) \\
    &V_\pi(\mathbf{s}_t) = \mathbb{E}_{\mathbf{a}_t\sim \pi}[Q_\pi(\mathbf{s}_t,\mathbf{a}_t) - \alpha\ln\pi(\mathbf{a}_t\mid\mathbf{s}_t)],
\end{aligned}
\end{equation}
where the Q function $Q_\pi$ evaluates the control policy, and $\alpha$ denotes the entropy regularization coefficient, often referred to as \emph{Temperature}~\cite{haarnoja2018composable}. The choice of $\alpha$ determines the relative importance of the entropy term against the reward, and a higher $\alpha$ corresponds to a larger preference for exploration to prevent premature convergence to suboptimal policies. We leave out the mathematical derivations for simplicity, and the details can be found in~\cite{haarnoja2018soft, haarnoja2019Soft, xuezhi2023optimal}.

\begin{figure}[!htb]
    \centering
    \includegraphics[width=3in]{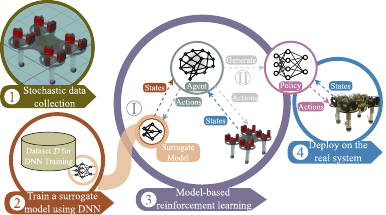}
    \caption{Gait control policy generation framework.}
    \label{fig:exp}
\end{figure}

\subsection{Training Framework}
Figure~\ref{fig:exp} illustrates our framework for developing an effective gait control policy for SoftQ. Our previous work uses a SimScape Multibody dynamics model of SoftQ as the environment for RL training, which is time-consuming. To reduce the training time, a Deep Neural Network (DNN) is chosen to represent the surrogate model $\hat{f}_\theta(\mathbf{s}_t, \mathbf{a}_t)$. The model is trained on a dataset $\mathcal{D}$ collected from the high-fidelity SimScape simulation model, where the dataset is collected by applying stochastic actions to the SimScape model. It encapsulates the transitions of current states $\mathbf{s}_t$ with actions $\mathbf{a}_t$ to next states $\mathbf{s}_{t+1}$. To evaluate the accuracy of the surrogate model, we utilize a dedicated dataset $\mathcal{D}_\textrm{val}$ that includes expertly designed gait patterns serving as a benchmark. Both datasets, $\mathcal{D}$ and $\mathcal{D}_\textrm{val}$, include pairs of current observations $\mathbf{s}_t$ and current actions $\mathbf{a}_t$ with their corresponding states $\mathbf{s}_{t+1}$ observed.

In the subsequent step, a surrogate model using DNN is trained to approximate the physics-based robot simulator. Building upon the surrogate models from the previous phases, we proceeded to \raisebox{.5pt}{\textcircled{\raisebox{-.9pt} {\uppercase\expandafter{\romannumeral1}}}} MBRL: train the control policy, and \raisebox{.5pt}{\textcircled{\raisebox{-.9pt} {\uppercase\expandafter{\romannumeral2}}}} PT: continuously refine the policy through SimScape simulation for higher accuracy. This continuous refinement is necessary to adapt the learned policy to the physical system. Finally, the trained control policy is implemented on the SoftQ to validate its effectiveness.  

\subsection{Robot Specifications}
\begin{figure}[!ht]
	\centering
		\includegraphics[width=3in]{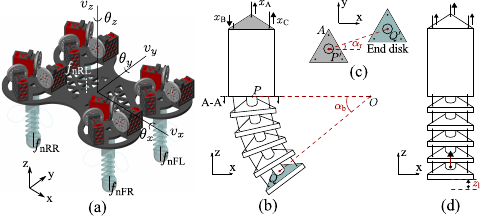}
	\caption{Overview of SoftQ and CTSA: (a) Rendered robot with key states. (b) CTSA bending angle $\alpha_b$. (c) CTSA rotational angle $\alpha_r$. (d) CTSA compression length $z_\textrm{l}$.} 
    \label{fig:robotsmodel}
\end{figure}

Our previous studies~\cite{ji2022omnidirectional, ji2022synthesizing} have developed SoftQ, a soft quadruped robot characterized by its innovative utilization of four CTSAs. Figure~\ref{fig:robotsmodel}(a) provides an illustration of the robot, including orientation (roll $\theta_x$, pitch $\theta_y$, and yaw $\theta_z$), translational velocities along three axes ($v_x, v_y, v_z$), and contact forces on its four feet ($f_\textrm{nFL}, f_\textrm{nFR}, f_\textrm{nRL}, f_\textrm{nRR}$). The defining characteristic of the CTSAs is the cable-driven mechanism, which can continuously bend and twist. The cables, also referred to as tendons, guide the deformation of soft materials. Each leg of the robot is actuated by three cables, driven by three servo motors, totaling 12 actuators for the robot's motion. Then the control input to each leg is a vector $\mathbf{d}_\textrm{ten} = \left[d_\textrm{A}, d_\textrm{B}, d_\textrm{C}\right]^\top$, as illustrated in Figure~\ref{fig:robotsmodel}(b). 

\section{Surrogate Model Development}
\label{sec:data_driven}

\subsection{Inverse Kinematics Analysis}

Based on the previous work~\cite{ji2022omnidirectional}, a leg's pose during the walking is characterized by three values illustrated in Figure~\ref{fig:robotsmodel}(b)-(d). $\alpha_\textrm{b}$ is the bending angle of a leg, $\alpha_\textrm{r}$ is the rotational angle of a leg, and $z_\textrm{l}$ is the compression length of a leg. A combination of the three values defines the pose of the leg $\mathbf{a} = [\alpha_\textrm{b}, \alpha_\textrm{r}, z_\textrm{l}]^\top$. The pose and the movements of the three servo motors satisfy the kinematic model with
\begin{equation}
    \label{eq:inverse_kinematic}
    \mathbf{d}_\textrm{ten} = g_\textrm{inv}(\mathbf{a}) = f_\textrm{inv}([\alpha_\textrm{b}, \alpha_\textrm{r}]^\top) + \textbf{1} z_\textrm{l},
\end{equation}
where $f_\textrm{inv}: \mathbb{R}^2 \rightarrow \mathbb{R}^3$ is derived by geometric analysis, and $g_\textrm{inv}(\cdot)$ is expressed by
\begin{equation}
\label{eq:geometric_analysis}
\begin{aligned}
    &d_\textrm{A} = R_\textrm{d} \alpha_\textrm{b} \cos(\alpha_\textrm{r})+z_l,\\
    &d_\textrm{B} = R_\textrm{d} \alpha_\textrm{b} \cos(\alpha_\textrm{r}+\frac{2\pi}{3})+z_l,\\
    &d_\textrm{C} = R_\textrm{d} \alpha_\textrm{b} \cos(\alpha_\textrm{r}+\frac{4\pi}{3})+z_l.
\end{aligned}
\end{equation}

The vector of tendon displacement $\mathbf{d}_\textrm{ten}$ is further used as the motor reference for the low-level position control. The synchronization of the designed trajectory on $\mathbf{a}$ and $\mathbf{d}_\textrm{ten}$ of each leg will lead to a gait pattern of the robot.

\subsection{Restriction on State Space}
\label{sec:restrict}

Built upon the inverse kinematic model in~\eqref{eq:inverse_kinematic} and~\eqref{eq:geometric_analysis}, we impose constraints on actions based on gait patterns and encapsulate the gait by parameters to increase learning efficiency. Defining specific gait patterns for rigid quadruped robots, such as trot, pace, bound, pronk, and gallop has been a prevalent practice~\cite{biswal2021Development}. By restricting gait controllers based on these patterns, we can simplify the controllers using specific gaits. The simplification reduces the search space and thus improves learning efficiency. Our prior approach~\cite{ji2022synthesizing} is built on the motors' behaviors, in particular for three motors to each leg, resulting in 12 actions, i.e., $\mathbf{a}_{t}^\textrm{MF}\!\in\!\mathbb{R}^{12}$. 
In this study, the robot is restricted to following the trot gait so that a pair of diagonal legs must move synchronously. The trot gait, noted for its stability and balance~\cite{vukobratovic1972Stability}, involves a two-beat diagonal movement pattern where diagonal leg pairs move synchronously~\cite{hyun2014Higha}. 
Consequently, the action space is restricted to the dynamics of diagonal leg pairs, defined by the desired bending angles ($\alpha_{\textrm{b}_\textrm{1}}\,,\alpha_{\textrm{b}_\textrm{2}}$) and compressed leg length ($z_{\textrm{l}_\textrm{1}}\,,z_{\textrm{l}_\textrm{2}}$) for each pair, resulting in an action space $$\mathbf{a}_t = [\alpha_{\textrm{b}_\textrm{1}}, z_{\textrm{l}_\textrm{1}}, \alpha_{\textrm{b}_\textrm{2}}, z_{\textrm{l}_\textrm{2}}]\:\in\:\mathbb{R}^{4}$$ The subscript 1 represents the pair of front left and rear right legs, and the subscript 2 represents the other pair. This reduction in action space is significant, bringing the space down from 10 to 4. Note that $\alpha_r$ is excluded from the action space due to its direct correlation with the robot's walking direction~\cite{ji2022omnidirectional}.

The state variables of the robot include sensor measurements from the quadruped robot, with orientation $\pmb{\theta}(t)=[\theta_x(t),\theta_y(t),\theta_z(t)]\in\mathbb{R}^3$, capturing roll, pitch, and yaw angles; translational velocities $\mathbf{v}(t)=[v_x(t),v_y(t),v_z(t)]\in\mathbb{R}^3$; and normalized contact forces $\mathbf{f}_\textrm{n}(t)=[f_\textrm{nFL}(t), f_\textrm{nFR}(t), f_\textrm{nRR}(t), f_\textrm{nRL}(t)]\in\mathbb{R}^4$. Thus the observation state vector to describe the robot is $$\mathbf{s}_t=[\pmb{\theta}(t),\mathbf{v}(t),\mathbf{f}_\textrm{n}(t)]\in\mathbb{R}^{10}$$

\begin{figure}[!ht]
    \centering
    \includegraphics[width=3in]{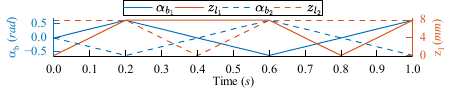}
    \caption{Expert gait design, solid lines for FL and RR pairs, dashed lines for FR and RL pairs.}
    \label{fig:expert}
\end{figure}

This study also leverages an expert trot gait from previous studies~\cite{ji2022omnidirectional} with a period of $T_\textrm{exp}$ = 0.8~$s$. A pair of diagonal legs moves identically, and the movements of the two pairs have a time delay of $T_\textrm{exp}/2$. When a leg swings backward from a forwarding bending pose, it is not compressed and thus has contact with the ground. When the leg swings forward, it is compressed to avoid touching the ground. Thus, the actions of the expert trot gait can be expressed in $\alpha_\textrm{b}$ and $z_\textrm{l}$, as plotted in Figure~\ref{fig:expert}. The diagonal leg pair only touches the ground during backward motion, relying on ground friction for forward propulsion. The further visual representation and analysis of the resulting trajectory are shown in Figure~\ref{fig:gait}.

\subsection{Surrogate Model Training}
\label{sec:surrogate}
The dynamics of the SimScape environment model is modeled by the state transition function $f$: $\mathcal{S} \times \mathcal{A} \rightarrow \mathcal{S}$, where $\mathcal{S}\in\mathbb{R}^{10}$ and $\mathcal{A}\in\mathbb{R}^{4}$, and the sampling time is $T_s$. The state update of this function relies on the execution of the SimScape model and is time-consuming. To reduce the computation time, we approximate $f$ with a surrogate model $\hat{f}_\theta: \mathcal{S}\times \mathcal{A} \rightarrow \mathcal{S}$, where $\theta$ is the vector of all parameters of the surrogate model. 
The training dataset $\mathcal{D}\subset \mathcal{S} \times \mathcal{A} \times \mathcal{S}$ is collected from executing the SimScape model with trajectories of random actions from a fixed initial state vector $\mathbf{s}_0$. Specifically, a certain percentage (2\%) of expert gait trajectories is injected into the dataset to improve the chances of successful walk experiences. The training of the surrogate model $\hat{f}_\theta$ finds the optimal model parameter $\theta$ that minimizes the average error for the dataset $\mathcal{D}$ by
\begin{equation}
    \text{Error} = \frac{1}{|\mathcal{D}|}\sum_{(\mathbf{s}_t,\mathbf{a}_t, \mathbf{s}_{t+1}) \in \mathcal{D}} \frac{1}{2}\lVert \mathbf{s}_{t+1}-\hat{f}_\theta(\mathbf{s}_t, \mathbf{a}_t)\rVert^2,
\label{eq:train_loss}
\end{equation}
where $\lVert\cdot\rVert$ denotes the Euclidean norm. 

We employ two metrics to quantify the accuracy of the surrogate model: the correlation coefficient (R) and the normalized root mean squared error (NRMSE). And
\begin{equation}
    \textrm{R}=\rho_\textrm{val} = \frac{\sum_{\mathcal{D}_{\textrm{val}},\hat{\mathcal{D}}_{\textrm{val}}}(\mathbf{s}_t - \Bar{\mathbf{s}})(\hat{\mathbf{s}}_t-\Bar{\mathbf{s}}_\textrm{pred})}{\sqrt{\sum_{\mathcal{D}_{\textrm{val}}}(\mathbf{s}_t - \Bar{\mathbf{s}})^2 \sum_{\mathcal{D}_{\textrm{val}}}(\hat{\mathbf{s}}_t-\Bar{\mathbf{s}}_\textrm{pred})^2}},
    \label{eq:R}
\end{equation}
where $\mathbf{s}_t$ is the measured state in validation dataset $\mathcal{D}_{\textrm{val}}$. $\hat{\mathbf{s}}_t\in\hat{\mathcal{D}}_{\textrm{val}}$ is the corresponding prediction states by the surrogate model. Similarly, $\bar{s} = \frac{1}{|\mathcal{D}_{\textrm{val}}|}\sum_{\mathcal{D}_{\textrm{val}}} \mathbf{s}_t$ is the average of all states in $\mathcal{D}_{\textrm{val}}$ and $\Bar{\mathbf{s}}_\textrm{pred} = \frac{1}{|\hat{\mathcal{D}}_{\textrm{val}}|}\sum_{\hat{\mathcal{D}}_{\textrm{val}}}\hat{\mathbf{s}}_t$ is the average of all prediction states. NRMSE is defined by
\begin{equation}
    \textrm{NRMSE} = \sqrt{\frac{1}{|\mathcal{D}_{\textrm{val}}|}\sum_{(\mathbf{s}_t, \mathbf{a}_t, \mathbf{s}_{t+1}) \in \mathcal{D}_{\textrm{val}}} \lVert \frac{\mathbf{s}_{t+1}-\hat{f}_\theta(\mathbf{s}_t, \mathbf{a}_t)}{\textrm{max}(\mathbf{s}_t) - \textrm{min}(\mathbf{s}_t)}\rVert^2}.
    \label{eq:NRMSE}
\end{equation}

Eqs.~\eqref{eq:R} and \eqref{eq:NRMSE} evaluate the prediction accuracy for one-step prediction. Nevertheless, the prediction accuracy in a multi-step lookahead scheme~\cite{bertsekas2019reinforcement} is also essential for the surrogate model to pursue global optimality.
We hence calculate $T$-step validation errors by propagating the learned dynamics function forward $T$ times to make multi-step open loop predictions. For each given sequence of true actions ($\mathbf{a}_t,...,\mathbf{a}_{t+T}$) from $\mathcal{D}_\textrm{val}$, we compare the corresponding ground-truth states ($\mathbf{s}_{t+1},...,\mathbf{s}_{t+T+1}$) to the surrogate model’s multi-step state predictions ($\hat{\mathbf{s}}_{t+1},...,\hat{\mathbf{s}}_{t+T+1}$) as follows. Let the initial state prediction $\hat{\mathbf{s}}_{t}$ be $\mathbf{s}_t$, the iterative states become $\hat{\mathbf{s}}_{t+i} = \hat{f}_\theta(\hat{\mathbf{s}}_{t+i-1}, \mathbf{a}_{t+i-1})$ for $i>0$, and
\begin{equation}
\text{NRMSE}^{T} = \sqrt{\frac{1}{|\mathcal{D}_\textrm{val}|}\sum_{\mathbf{s}_t \in \mathcal{D}_\textrm{val}}\frac{1}{T}\sum_{i=1}^{T}\Big\lVert \frac{\mathbf{s}_{t+i}-\hat{\mathbf{s}}_{t+i}}{\textrm{max}(\mathbf{s}_t) - \textrm{min}(\mathbf{s}_t)} \Big\rVert^2}.
\label{eq:NRMSET}
\end{equation}
Similarly, the R in $T$-step validation can be updated with
\begin{equation}
\text{R}^{T} = \frac{1}{T}\sum_{i=1}^{T}\rho_{\textrm{val}, i}.
\label{eq:RT}
\end{equation}

\begin{figure}[!ht]
    \centering
    \includegraphics[width=3in]{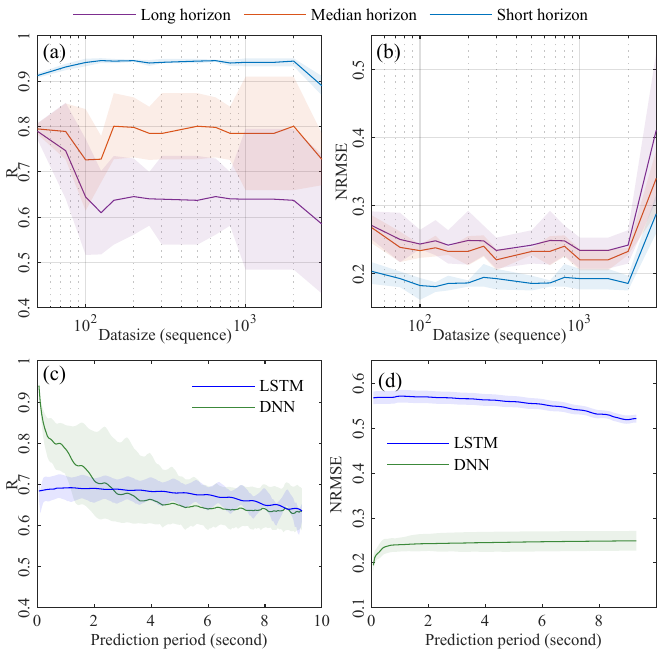}
    \caption{Evaluation of the surrogate model accuracy with varying training data sizes of DNN: (a) R and (b) NRMSE as a function of dataset size. Prediction performance of two architectures at the selected dataset size (250) in terms of (c) $\text{R}^{T}$ and (d) $\text{NRMSE}^{T}$.}
    \label{fig:size_t}
\end{figure}

The accuracy of the surrogate model relies heavily on the dataset size~\cite{ibarz2021How}. In this study, the dataset size is quantified by the number of trajectory sequences, rather than individual samples, aiming to capture the holistic nature of continuous walking. It also ensures that the surrogate model generalizes effectively across various phases and intricacies inherent in consecutive steps. This approach contributes to a more robust and applicable model for gait training in MBRL. The relationship between the data size and the accuracy of the surrogate model is analyzed in Figure~\ref{fig:size_t}. When the data size is less than 200 sequences, the short horizon prediction accuracy improves when the step size increases, because the corresponding R value increases and NRMSE decreases. The prediction accuracy for multiple-step prediction, however, drops when the data size increases, as R value decreases and NRMSE reaches saturation, suggesting a limit to prediction accuracy within the normalized value range.

We studied both Long Short-Term Memory (LSTM) and DNN for representing the surrogate model. As an evolution of RNNs, LSTM layers use specialized memory cells and gating mechanisms to selectively retain or forget information over extended sequences. In the SoftQ surrogate model, a bidirectional LSTM layer is adopted, involving two distinct LSTM layers. The LSTM architecture includes a fully connected layer, a bidirectional LSTM layer, and a random dropout layer of 50\%, as shown in TABLE~\ref{tab:structure}. The DNN architecture comprises three hidden layers, also as shown in TABLE~\ref{tab:structure}.

As illustrated in Figure~\ref{fig:size_t}(a)-(b), when the data size is larger than 200 but less than 2000, the R and NRMSE values for one-step and multiple-step predictions are stable. The prediction accuracy becomes worse when the data size is larger than 2000. The reason can be attributed to factors such as an imbalanced data distribution and limited representation of critical information regarding the robot's dynamics, particularly in scenarios with high walking velocities. 

Based on the above analysis, we choose the data size of 250 sequences for learning the DNN and LSTM surrogate models. The prediction accuracy of the learned model is illustrated in Figure~\ref{fig:size_t}(c)-(d). Evidently, the prediction accuracy decreases with a higher prediction length. However, NRMSE does not exhibit oscillations but instead begins to increase for the first few time horizons ($T$$\leq$16$T_s$), ultimately reaching a saturation point at around a 20-step horizon prediction ($T$=1~$s$). This behavior aligns with the gait period of the validation gait (i.e., 0.8~$s$). In addition, the DNN surrogate model has higher one-step prediction accuracy than the LSTM surrogate model, because it has higher R and smaller NRMSE when the prediction period is 1~$s$. When the prediction period increases, the accuracy of both surrogate models decreases. When the prediction period reaches 10~$s$, the R values of the two models are very close and the NRMSE value of the DNN model is much less than that of the LSTM model. The quantitative comparison shows that the DNN model is more accurate than the LSTM model. Therefore, the DNN network is selected for the surrogate model.

\begin{table}[!ht]
    \centering
    \caption{Structures of Neural Networks}
    \label{tab:structure}
    \begin{adjustbox}{width=\linewidth}
\begin{tabular}{c|ccc}
Network   & In \& Out  & Hidden Layers & Activation \\ \hline
Surrogate & Feature(14) $\rightarrow$ & Fully Connected & ReLU \\
(DNN)     & Regression & (64, 128, 64)  &   \\
Surrogate & Feature(14) $\rightarrow$ & Fully Connected(64)$\rightarrow$ & ReLU \\
(LSTM)    & Fully Connected & LSTM(100)$\rightarrow$ Dropout(0.5) &  \\
          & Feature(14) $\rightarrow$ & Fully Connected & ReLU(32) \\
Critic    & Fully Connected & (128, 128, 128) &  \\
          & Feature(14) $\rightarrow$ & Fully Connected & ReLU(4) + \\
Actor     & Concatenation &  (256, 128, 128) & Softplus(4)     
\end{tabular}
\end{adjustbox}
\end{table}

\begin{table}[!ht]
    \centering
    \caption{Hyper-parameters for RL training}
    \label{tab:hyperpara}
    \begin{adjustbox}{width=\linewidth}
    \begin{tabular}{c|cccc}
       Parameters & MBRL & MFRL & Post training & Ji et al.~\cite{ji2022synthesizing} \\\hline
       MaxEpi & 400 & 600 & 400 & 1500\\
       MaxStep & 100 & 100 & 100 & 100\\
       $H'$ & -4 & -4 & -4 & -12\\
       $\alpha_c$ & 0.002 & 0.002 & 0.001 & 0.002\\
       $\alpha_a$ & 0.001 & 0.001 & 0.0005 & 0.001\\
       $\alpha$ & 0.001 & 0.001 & 0.001 & 0.001\\
       batch size & 4096 ($2^{12}$) & 4096 & 4096 & 512\\
       replay buffer & 16384 ($2^{14}$) & 16384  & 16384 & 4096\\
    \end{tabular}
    \end{adjustbox}
\end{table}

\section{Model Based Reinforcement Learning}
\label{sec:MBRL}
In this study, SAC utilizes two Q-value critic networks alongside their respective target critic networks to estimate the Q function. The structures of both networks are summarized in TABLE~\ref{tab:structure}. Both critics have an identical network structure comprising two input sections for observations and actions. The observation component consists of two fully-connected layers, each containing 128 units, while the action component is processed through a single fully-connected layer with 128 units. The outputs from these components are concatenated and further processed by a ReLU-activated fully-connected layer with 32 units, yielding the Q network's output. On the other hand, the actor network takes environmental observations, subjects them to two ReLU-activated fully-connected layers, each with 256 units, and predicts action means and variances utilizing Gaussian distributions. The resulting actions are constrained within the range [0, 1] using the hyperbolic tangent function (tanh).

The hyper-parameters for the RL algorithms are summarized in TABLE~\ref{tab:hyperpara}, following the configuration outlined in the prior work~\cite{ji2022synthesizing}. To mitigate overfitting, we set the learning rates for the critic $\alpha_c$ and actor $\alpha_a$ networks to 0.002 and 0.001 both in MBRL and MFRL, respectively. Additionally, all the temperature parameter $\alpha$ shares a learning rate of 0.001. Policy network updates occur every 3 simulation steps for stability. The target entropy is derived from the number of actions, with $H'=-4$. Episodes begin from the same initial state $\mathbf{s}_0$ to ensure consistency in environment initialization. To enhance the robustness of the trained policy, artificial noise is introduced to both the SimScape and the surrogate models during the training process. This noise, based on sensor calibration results, is applied to velocity, angular, and force signal data with variances set to $\sigma^2_{\mathbf{v}} = 0.002$, $\sigma^2_{\pmb{\theta}} = 0.002$, and $\sigma^2_{\mathbf{f}_n} = 0.005$, respectively.

\subsection{Agent Specifications and Reward}
To account for the slow dynamics of the soft legs and potential non-Markovian property in the physical deployment~\cite{ibarz2021How}, the previous action $\mathbf{a}_{t-1} \in \mathbb{R}^4$ is added to the robot's state vector, and the full state vector in the RL training is appended to $\mathbf{s}_t \leftarrow [\mathbf{s}_t, \mathbf{a}_{t-1}]\:\in\:\mathbb{R}^{14}$. Thus, the refined input state representation encompasses 14 key variables $\mathbf{s}_t\!=\![\pmb{\theta}(t),\mathbf{v}(t),\mathbf{f}_n(t),\mathbf{a}_{t-1}]\!\in\!\mathbb{R}^{14}$, which include the robot's orientation, velocities, normalized contact forces, and the action executed at the previous time step.
Note that this refinement is exclusively intended for RL training purposes and does not impact the definition of the surrogate model's state space in Sec.~\ref{sec:surrogate}.
Actions are defined in terms of desired bending angles ($\alpha_{\textrm{b}_\textrm{1}}\,,\alpha_{\textrm{b}_\textrm{2}}$) and compressed lengths ($z_{\textrm{l}_\textrm{1}}\,,z_{\textrm{l}_\textrm{2}}$) for the two pairs of legs, where subscript 1 represents the pair of front left and rear right legs, and subscript 2 represents the other pair. Actions are also normalized to [0, 1] to confine the policy network output, ensuring stability and convergence during training.

The reward function $r(\mathbf{s}_t, \mathbf{a}_t)$ is formulated to promote a linear and stable gait over time, where
\begin{equation}
    \begin{aligned}
    &r(\mathbf{s}_t,\mathbf{a}_t) = \epsilon_1\frac{T_s}{T_f} + (1-\epsilon_2|v_x(t)-v_\textrm{ref}|)\\
    &-\epsilon_3\lVert\Ddot{\mathbf{a}}_t\rVert - \epsilon_4\lVert\mathbf{a}_t-\mathbf{\sigma}_\textrm{threshold}\rVert - \epsilon_5\Big(\mathbf{a}_t-\frac{\sum_{i=1}^{T}\mathbf{a}_i}{T}\Big)^2.
    \end{aligned}
    \label{eq:reward}
\end{equation}

In \eqref{eq:reward}, the robot receives a constant reward $\epsilon_1\frac{T_s}{T_f}$ if it maintains balance for each training step. $T_s$ represents the sampling time and $T_f$ is the training time of one episode. The second term encourages the robot to follow a reference speed at $v_\textrm{ref}$. We set $\epsilon_2 = 1/v_\textrm{ref}$ for simplicity, ensuring a maximum reward of one.

To avoid jerky action, a penalty on large action acceleration ($\Ddot{\mathbf{a}}_t$) is introduced with $\epsilon_3$. This penalty discourages abrupt action changes using finite differences of actions from the last three time steps.

To ensure stable walking, excessive leg bending angles $\alpha_b$ are penalized. Negative rewards, i.e., $- \epsilon_4 \lVert\mathbf{a}_t-\sigma_\textrm{threshold}\rVert$, discourage excessive leg bending. An additional penalty term of $ - \epsilon_5(\mathbf{a}_t-\frac{\sum_{i=1}^{T}\mathbf{a}_i}{T})^2$ discourages consistent leg bending in the same direction over a horizon $T$. The weighting factors (i.e., $\pmb{\epsilon} = [\epsilon_1, \epsilon_2, \epsilon_3, \epsilon_4, \epsilon_5]$) define the reward function's emphasis, set as $[5, 1/v_\textrm{ref}, 0.25, 10, 3]$ to align with effective robot training.

\subsection{Post-training}
To compensate for the error of the surrogate model, we employ PT to refine the control policy. Note that model-reliant controllers, e.g., MPC or Monte Carlo Tree Search (MCTS), are not readily employed for the robot control, mainly due to potential sub-optimal solutions resulting from the inaccuracies in the surrogate model.

In the PT phase, the learning rates are adjusted to 0.001 for the critic networks and 0.0005 for the actor network, which are half of the values used in MBRL. This reduction is implemented to moderate the learning process during PT, preventing rapid adjustments that may destabilize the already well-established control policy developed in the earlier MBRL training.

To accelerate the training process with the SimScape model, a predefined expert control policy with 1.6~$s$ is introduced at the beginning of each training episode. This approach is similar to \emph{imitation learning}~\cite{shao2022Learning}, although this expert gait does not ensure optimal performance. Instead, the defined expert policy allows the RL agent to quickly adjust its behavior for efficient real-time implementation.


We maintain the same RL hyper-parameters with adjustments for optimizing gait speed in PT, setting $v_\textrm{ref}$ at 0.3~$m/s$. Reward function coefficients are adapted to $[\epsilon_1, \epsilon_2, \epsilon_3, \epsilon_4, \epsilon_5] = [5, 1/v_\textrm{ref}, 0.25, 100, 3]$ to avoid excessive bending and compression of the legs.

\section{Results and Validation}
\label{sec:result_validation}

\subsection{Training Performance}
\begin{figure}[!ht]
    \centering
    \includegraphics[width=3in]{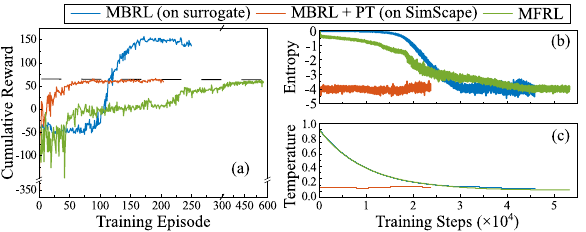}
    \caption{The training results in 0.2 $m/s$ reference speed. (a) Cumulative reward with training episodes. Variations in (b) entropy and (c) temperature during the training process.}
    \label{fig:train}
\end{figure}

\begin{figure}[!ht]
    \centering
    \includegraphics[width=3in]{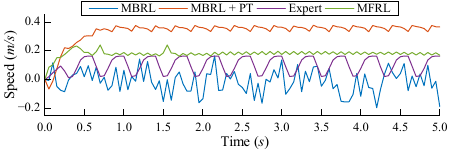}
    \caption{Resultant forward walking speed in simulation for expert gait, MBRL, MBRL+PT (fastest), and MFRL (fastest).}
    \label{fig:simu}
\end{figure}

The training curves of MBRL using the DNN surrogate model is visualized in Figure~\ref{fig:train}, and the cumulative reward converges to a high value at around 180 within 200 episodes of 3.5 hours. However, when the learned control policy is applied to the SimScape environment model of the robot, the resulting walking speed is much less than the result with the surrogate model, as indicated in Figure~\ref{fig:simu}.

In Figure~\ref{fig:train}, we also present the training outcomes for both MBRL with post-training (MBRL+PT) and MFRL using the same SimScape model. MBRL+PT exhibits rapid enhancement, achieving convergence in cumulative reward within approximately 60 episodes. Conversely, MFRL reaches convergence after approximately 450 episodes. These findings suggest that MBRL+PT outperforms MFRL, as MFRL can only learn a walking gait with the extended training effort after around 600 episodes, as reported in the prior work~\cite{ji2022synthesizing}. 
In simulations, the robot efficiently exceeds the reference speed, covering a distance of 1.8~$m$ in 5~$s$ at an average speed of 0.36~$m/s$, further confirming the effectiveness of MBRL+PT. Compared with the expert gait ($\Bar{v}_x$ = 0.09~$m/s$) as shown in Figure~\ref{fig:simu}, MBRL+PT achieves even higher velocities. 

\subsection{Benchmark Comparison}
Multiple metrics are defined to verify the performance of the learned control policy. A stability metric is a weighted combination of gait duration, angular velocity on the $z$ axis ($\Dot{\theta}_z$), and velocity on the $y$ axis, given as 
\begin{equation}
\textrm{stability} = w_\textrm{time} t - w_{\Dot{\theta}}\textrm{max}(|\Dot{\theta}_z|) - w_{v}\textrm{max}(|v_y|),
\label{eq:stability}
\end{equation}
where $[w_\textrm{time},w_{\Dot{\theta}},w_{v}] = [0.2,1,1]$ are chosen to emphasize the significance of angular velocity ($\Dot{\theta}_z$) and velocity in the $y$ axis ($v_y$) in assessing the stability of the robot. $t$ is the time that the robot walks before failure and this stability is evaluated for the full episode. The stability metric is chosen and assessed according to the reward definition in~(\ref{eq:reward}), aiming to punish early fall downs and jerky motions.

Additionally, the average speed and the Cost of Transport (COT) are used for performance metrics. COT measures the robot's energy efficiency and is defined by $\textrm{COT} = \frac{E}{md}$, where $E$ represents the energy consumed, $m$ the mass of the robot, and $d$ the distance traveled. Furthermore, learning efficiency is evaluated based on the time taken by the algorithm to converge to a satisfactory policy and the cumulative reward obtained.

A series of training sessions are conducted across varied reference velocities ($v_\textrm{ref}$) of 0.2~$m/s$, 0.3~$m/s$, and 0.5~$m/s$. Each velocity condition was independently trained three times. Subsequent evaluations involved a 5-second simulation for each trained agent, and the performance outcomes are summarized in TABLE~\ref{table:comp_benchmark}.
\begin{table}[!ht]
    \centering
    \caption{Comparison of the performance}
    \label{table:comp_benchmark}
    \begin{adjustbox}{width=\linewidth}
        \begin{tabular}{r|cccc} 
            \multicolumn{1}{l}{} &     & MBRL+PT & MFRL & Expert \\ \hline
            \multirow{2}{*}{stability} & [min,max] & [-0.04, 0.8] &[0.02, 0.745] & - \\ 
                      & avg & 0.65 & 0.03 & 0.73  \\ \hline
            COT       & [min,max] & [41, 146] &[40, 1375]& - \\ 
            ($J/kg/m$)  & avg & 68 & 501 & 81 \\\hline
            training time& [min,max] & [7.9, 19.8] &[19.7, 61.8] & - \\
            ($h$)     & avg &  10.8  &  22.6 &  -  \\\hline
            speed     & [min,max] & [0.04, 0.47] &[0.002, 0.22] & - \\
            ($m/s$)     & avg & 0.26 & 0.003 & 0.09 \\
        \end{tabular}
    \end{adjustbox}
\end{table}

As shown in TABLE~\ref{table:comp_benchmark}, MBRL+PT exhibits a significantly higher average stability score, approximately 0.65, compared to MFRL's score of 0.03. In terms of COT, MBRL+PT approaches the energy efficiency of the expert gait, while MFRL has a much lower efficiency. Additionally, MBRL+PT also enjoys higher training efficiency, achieving effective locomotion in less time (10.8 hours on average excluding data collection) compared to MFRL, which required more than 600 episodes or 24 hours of training. MBRL+PT outperformed MFRL in terms of walking speed, with only one MFRL agent achieving walking after 400 episodes but requiring 61.8 hours of training. These results emphasize the effectiveness of MBRL+PT in achieving stable, efficient, and adaptive gait control, surpassing MFRL in multiple performance metrics.

\begin{figure}[!htb]
    \centering
    \includegraphics[width=3in]{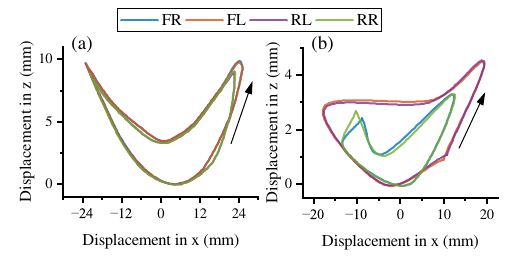}
    \caption{Foot trajectories for SoftQ by using (a) expert gait and (b) MBRL+PT.}
    \label{fig:gait}
\end{figure}

The trajectories of four feet achieved by the MBRL+PT controller and the expert gait are compared in Figure~\ref{fig:gait}. The MBRL+PT controller achieves smaller swing distances in the moving direction and in the $z$ axis compared to the expert gait. Thus, the MBRL+PT method achieves reduced swing area and a higher swing frequency ($T_\textrm{MB}$ = 0.3 $s$ vs. $T_\textrm{exp}$ = 0.8 $s$) for faster walking speeds. Additionally, synchronized motions are observed between the FR and RL leg pairs, as well as between the FL and RR leg pairs, reflecting paired movements. Notably, the FL and RR leg pairs exhibit a smaller swing area than the FR and RL pairs, possibly due to corrective actions addressing yaw rotation errors arising from a slightly larger mass distribution at the right side.

\subsection{Real Implementation: Control Architecture Design}
\begin{figure}[!htb]
	\centering
		\includegraphics[width=3in]{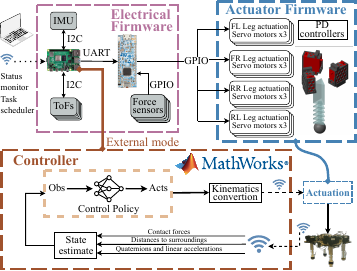}
	\caption{Control architecture.} \label{fig:architecture}
\end{figure}
The learning and control system architecture for SoftQ is illustrated in Figure~\ref{fig:architecture}. The PC in Figure~\ref{fig:architecture}, equipped with an Intel(R) Core(TM) i7-4770 CPU @3.40 GHz, 16 GB of memory, serves as both a MATLAB Simulink program transmitter and GUI for the robot operator. The Simulink program runs on a Raspberry Pi via external mode, gathering real-time data from ToF sensors and a lower-level controller.

Additionally, a NUCLEO-G431KB micro-controller board manages motors and reads analog data from force sensors. All controllers operate with a sampling time of $T_s = 0.05~s$ and further details of the experimental setup can be found in~\cite{ji2022synthesizing, xuezhi2023optimal}. Each leg is controlled by three servo motors and connected tendons, regulated by a PD controller to reach target positions assigned by the RL controller. Displacement speed components in the $x$, $y$, and $z$ axes are estimated via the integration of accelerations from IMU signals and ToF distance. Contact forces at leg ends are measured by force sensors. Reference signals are transmitted to servo motors from the PC via TCP communication over Wi-Fi. Sensor data from the robot is transmitted back to the PC via the same network used for reference signals reception.

\subsection{Physical Deployment Results}
\begin{figure}[!ht]
    \centering
    \includegraphics[width=3in]{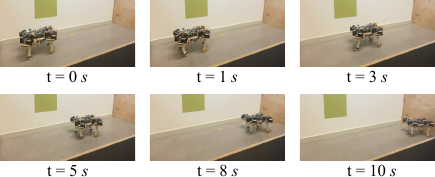}
    \caption{Field test results captured in video frames.}
    \label{fig:real_test}
\end{figure}

Furthermore, we deploy the obtained controller to assess its application in real-world scenarios. Two room dividers and walls serve both as navigational guides and as reference points for the robot's sensor array. The task is to let the robot traverse a distance of 1.5 $m$ (5 times of body length) and stop precisely 0.1 $m$ from the wall. The records in Figure~\ref{fig:real_test} show the robot's ability to effectively replicate the trained controller. The reference speed for the MBRL controller is 0.3 $m/s$, but the actual robot reaches an average speed of 0.15 $m/s$.

\begin{figure}[!ht]
    \centering
    \includegraphics[width=3in]{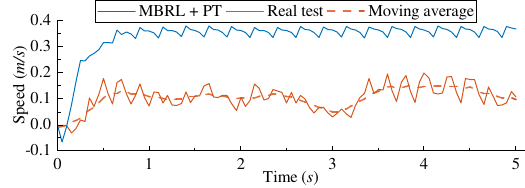}
    \caption{Comparison of speeds in real test and simulation.}
    \label{fig:real_vx}
\end{figure}

To address the sensor noise and the sim-to-real gap, we adjust sensor inputs by representing contact force as binary ($1$ when the foot touches the ground, $0$ otherwise) during the PT phase and implement a Kalman filter for velocity estimation in physical tests using ToF distance and IMU acceleration data. We set process noise covariance matrix $Q = [\begin{smallmatrix}0.01 & 0 \\ 0 & 1\end{smallmatrix}]$ and observation noise covariance matrix $R = 0.1$. Initial state estimates ($\mathbf{x}_0$) and initial error covariance estimates ($P_0$) are $\mathbf{x}_0 = [\begin{smallmatrix} 0 \\ 0 \end{smallmatrix}$] and $P_0 = [\begin{smallmatrix}1 & 0 \\ 0 & 1\end{smallmatrix}]$, respectively, to improve accuracy in real-world scenarios.

Figure~\ref{fig:real_vx} compares the robot's walking speed in simulation and physical experiments. The result exhibits the ideal performance with a forward speed of 0.36 $m/s$ and the robot walks at an average of 0.13 $m/s$ in reality. Notably, our approach outperformed the previous MFRL benchmark~\cite{ji2022synthesizing}, which achieved a speed of 0.05 $m/s$ in physical tests. However, there is a reduction in performance when transitioning from simulation to real-world experimentation. This discrepancy underscores the challenges inherent in transferring capabilities from simulated models to physical hardware, encompassing variations in actuator performance and the intricacies of foot-ground interaction. Addressing this reality gap by implementing our learning process in real-time could be an improvement for future development.

\section{Conclusion and Future Work}
\label{sec:conclusion}
This study presents significant advancements in soft quadruped robot control using MBRL with PT. The focus is on enhancing real-time control strategies, robust training methodologies, and training efficiency. Key contributions include demonstrating the efficiency of using a surrogate DNN model to replace the time-consuming environment model for gait controller training in soft quadruped robots, implementing a parametric inverse kinematics model to reduce state and action spaces, and empirically validating the real-world effectiveness and adaptability of their gait control methodology. Future work may explore MBRL algorithms to adapt soft quadruped robots to diverse terrains, employ real-time learning mechanisms in dynamic environments, further reduce the simulation-reality gap, enhance real-world walking speed, and extend RL control strategies to broader robotic applications.

\section*{Declaration of competing interest}
The authors declare that they have no known competing financial interests or personal relationships that could have appeared to influence the work reported in this paper.

\section*{Declaration of generative AI and AI-assisted technologies in the writing process}
During the preparation of this work the author(s) used ChatGPT in order to improve readability and language of the work. After using this tool/service, the author(s) reviewed and edited the content as needed and take(s) full responsibility for the content of the publication.

\section*{Acknowledgements}
This research has been carried out as part of the TECoSA Vinnova Competence Center for Trustworthy Edge Computing Systems and Applications at KTH Royal Institute of Technology (www.tecosa.center.kth.se). Lei Feng is also partly funded by KTH XPRES. 

\section*{CRediT authorship contribution statement}
$\textbf{Xuezhi~Niu, Kaige~Tan:}$ Conceptualization, Methodology, Software, Validation, Formal analysis, Investigation, Writing Original Draft, Visualization.
$\textbf{Lei~Feng:}$ Supervision, Project administration, Funding acquisition, Writing-Review $\&$ Editing.

\bibliographystyle{elsarticle-num}
\bibliography{myfiles}

\begin{thebibliography}{10}
\expandafter\ifx\csname url\endcsname\relax
  \def\url#1{\texttt{#1}}\fi
\expandafter\ifx\csname urlprefix\endcsname\relax\def\urlprefix{URL }\fi
\expandafter\ifx\csname href\endcsname\relax
  \def\href#1#2{#2} \def\path#1{#1}\fi

\bibitem{choi2023learning}
S.~Choi, G.~Ji, J.~Park, H.~Kim, J.~Mun, J.~H. Lee, J.~Hwangbo, \href{https://www.science.org/doi/abs/10.1126/scirobotics.ade2256}{Learning quadrupedal locomotion on deformable terrain}, Science Robotics 8~(74) (2023) eade2256.
\newblock \href {http://dx.doi.org/10.1126/scirobotics.ade2256} {\path{doi:10.1126/scirobotics.ade2256}}.
\newline\urlprefix\url{https://www.science.org/doi/abs/10.1126/scirobotics.ade2256}

\bibitem{bledt2018MIT}
G.~Bledt, M.~J. Powell, B.~Katz, J.~Di~Carlo, P.~M. Wensing, S.~Kim, {MIT} {Cheetah} 3: {Design} and {Control} of a {Robust}, {Dynamic} {Quadruped} {Robot}, in: 2018 {IEEE}/{RSJ} {International} {Conference} on {Intelligent} {Robots} and {Systems} ({IROS}), 2018, pp. 2245--2252.
\newblock \href {http://dx.doi.org/10.1109/IROS.2018.8593885} {\path{doi:10.1109/IROS.2018.8593885}}.

\bibitem{hutter2016anymal}
M.~Hutter, C.~Gehring, D.~Jud, A.~Lauber, C.~D. Bellicoso, V.~Tsounis, J.~Hwangbo, K.~Bodie, P.~Fankhauser, M.~Bloesch, R.~Diethelm, S.~Bachmann, A.~Melzer, M.~Hoepflinger, {ANYmal} - a highly mobile and dynamic quadrupedal robot, in: 2016 {IEEE}/{RSJ} {International} {Conference} on {Intelligent} {Robots} and {Systems} ({IROS}), 2016, pp. 38--44.
\newblock \href {http://dx.doi.org/10.1109/IROS.2016.7758092} {\path{doi:10.1109/IROS.2016.7758092}}.

\bibitem{taheri2023Studya}
H.~Taheri, N.~Mozayani, \href{https://www.sciencedirect.com/science/article/pii/S0094114X23002197}{A study on quadruped mobile robots}, Mechanism and Machine Theory 190 (2023) 105448.
\newblock \href {http://dx.doi.org/10.1016/j.mechmachtheory.2023.105448} {\path{doi:10.1016/j.mechmachtheory.2023.105448}}.
\newline\urlprefix\url{https://www.sciencedirect.com/science/article/pii/S0094114X23002197}

\bibitem{yasa2023Overview}
O.~Yasa, Y.~Toshimitsu, M.~Y. Michelis, L.~S. Jones, M.~Filippi, T.~Buchner, R.~K. Katzschmann, \href{https://www.annualreviews.org/content/journals/10.1146/annurev-control-062322-100607}{An {Overview} of {Soft} {Robotics}}, Annual Review of Control, Robotics, and Autonomous Systems 6~(Volume 6, 2023) (2023) 1--29.
\newblock \href {http://dx.doi.org/10.1146/annurev-control-062322-100607} {\path{doi:10.1146/annurev-control-062322-100607}}.
\newline\urlprefix\url{https://www.annualreviews.org/content/journals/10.1146/annurev-control-062322-100607}

\bibitem{drotman20173d}
D.~Drotman, S.~Jadhav, M.~Karimi, P.~de~Zonia, M.~T. Tolley, {3D} printed soft actuators for a legged robot capable of navigating unstructured terrain, in: 2017 {IEEE} {International} {Conference} on {Robotics} and {Automation} ({ICRA}), 2017, pp. 5532--5538.
\newblock \href {http://dx.doi.org/10.1109/ICRA.2017.7989652} {\path{doi:10.1109/ICRA.2017.7989652}}.

\bibitem{ji2022omnidirectional}
Q.~Ji, S.~Fu, L.~Feng, G.~Andrikopoulos, X.~V. Wang, L.~Wang, Omnidirectional walking of a quadruped robot enabled by compressible tendon-driven soft actuators, in: 2022 {IEEE}/{RSJ} {International} {Conference} on {Intelligent} {Robots} and {Systems} ({IROS}), 2022, pp. 11015--11022.
\newblock \href {http://dx.doi.org/10.1109/IROS47612.2022.9981314} {\path{doi:10.1109/IROS47612.2022.9981314}}.

\bibitem{wang2022control}
J.~Wang, A.~Chortos, \href{https://onlinelibrary.wiley.com/doi/abs/10.1002/aisy.202100165}{Control {Strategies} for {Soft} {Robot} {Systems}}, Advanced Intelligent Systems 4~(5) (2022) 2100165.
\newblock \href {http://dx.doi.org/10.1002/aisy.202100165} {\path{doi:10.1002/aisy.202100165}}.
\newline\urlprefix\url{https://onlinelibrary.wiley.com/doi/abs/10.1002/aisy.202100165}

\bibitem{rus2015design}
D.~Rus, M.~T. Tolley, \href{https://www.nature.com/articles/nature14543}{Design, fabrication and control of soft robots}, Nature 521~(7553) (2015) 467--475.
\newblock \href {http://dx.doi.org/10.1038/nature14543} {\path{doi:10.1038/nature14543}}.
\newline\urlprefix\url{https://www.nature.com/articles/nature14543}

\bibitem{fahmi2019passive}
S.~Fahmi, C.~Mastalli, M.~Focchi, C.~Semini, \href{https://ieeexplore.ieee.org/abstract/document/8678400/authors#authors}{Passive {Whole}-{Body} {Control} for {Quadruped} {Robots}: {Experimental} {Validation} {Over} {Challenging} {Terrain}}, IEEE Robotics and Automation Letters 4~(3) (2019) 2553--2560.
\newblock \href {http://dx.doi.org/10.1109/LRA.2019.2908502} {\path{doi:10.1109/LRA.2019.2908502}}.
\newline\urlprefix\url{https://ieeexplore.ieee.org/abstract/document/8678400/authors#authors}

\bibitem{dudzik2020robust}
T.~Dudzik, M.~Chignoli, G.~Bledt, B.~Lim, A.~Miller, D.~Kim, S.~Kim, \href{https://ieeexplore.ieee.org/abstract/document/9340701}{Robust {Autonomous} {Navigation} of a {Small}-{Scale} {Quadruped} {Robot} in {Real}-{World} {Environments}}, in: 2020 {IEEE}/{RSJ} {International} {Conference} on {Intelligent} {Robots} and {Systems} ({IROS}), 2020, pp. 3664--3671.
\newblock \href {http://dx.doi.org/10.1109/IROS45743.2020.9340701} {\path{doi:10.1109/IROS45743.2020.9340701}}.
\newline\urlprefix\url{https://ieeexplore.ieee.org/abstract/document/9340701}

\bibitem{sleiman2021unified}
J.-P. Sleiman, F.~Farshidian, M.~V. Minniti, M.~Hutter, \href{https://ieeexplore.ieee.org/abstract/document/9387121}{A {Unified} {MPC} {Framework} for {Whole}-{Body} {Dynamic} {Locomotion} and {Manipulation}}, IEEE Robotics and Automation Letters 6~(3) (2021) 4688--4695.
\newblock \href {http://dx.doi.org/10.1109/LRA.2021.3068908} {\path{doi:10.1109/LRA.2021.3068908}}.
\newline\urlprefix\url{https://ieeexplore.ieee.org/abstract/document/9387121}

\bibitem{dicarlo2018Dynamic}
J.~Di~Carlo, P.~M. Wensing, B.~Katz, G.~Bledt, S.~Kim, Dynamic {Locomotion} in the {MIT} {Cheetah} 3 {Through} {Convex} {Model}-{Predictive} {Control}, in: 2018 {IEEE}/{RSJ} {International} {Conference} on {Intelligent} {Robots} and {Systems} ({IROS}), 2018, pp. 1--9.
\newblock \href {http://dx.doi.org/10.1109/IROS.2018.8594448} {\path{doi:10.1109/IROS.2018.8594448}}.

\bibitem{ponton2021efficient}
B.~Ponton, M.~Khadiv, A.~Meduri, L.~Righetti, \href{https://ieeexplore.ieee.org/abstract/document/9350175}{Efficient {Multicontact} {Pattern} {Generation} {With} {Sequential} {Convex} {Approximations} of the {Centroidal} {Dynamics}}, IEEE Transactions on Robotics 37~(5) (2021) 1661--1679.
\newblock \href {http://dx.doi.org/10.1109/TRO.2020.3048125} {\path{doi:10.1109/TRO.2020.3048125}}.
\newline\urlprefix\url{https://ieeexplore.ieee.org/abstract/document/9350175}

\bibitem{miki2022learning}
T.~Miki, J.~Lee, J.~Hwangbo, L.~Wellhausen, V.~Koltun, M.~Hutter, \href{https://www.science.org/doi/10.1126/scirobotics.abk2822}{Learning robust perceptive locomotion for quadrupedal robots in the wild}, Science Robotics 7~(62) (2022) eabk2822.
\newblock \href {http://dx.doi.org/10.1126/scirobotics.abk2822} {\path{doi:10.1126/scirobotics.abk2822}}.
\newline\urlprefix\url{https://www.science.org/doi/10.1126/scirobotics.abk2822}

\bibitem{tsounis2020deepgait}
V.~Tsounis, M.~Alge, J.~Lee, F.~Farshidian, M.~Hutter, \href{https://ieeexplore.ieee.org/abstract/document/9028188}{{DeepGait}: {Planning} and {Control} of {Quadrupedal} {Gaits} {Using} {Deep} {Reinforcement} {Learning}}, IEEE Robotics and Automation Letters 5~(2) (2020) 3699--3706.
\newblock \href {http://dx.doi.org/10.1109/LRA.2020.2979660} {\path{doi:10.1109/LRA.2020.2979660}}.
\newline\urlprefix\url{https://ieeexplore.ieee.org/abstract/document/9028188}

\bibitem{morimoto2021ModelFree}
R.~Morimoto, S.~Nishikawa, R.~Niiyama, Y.~Kuniyoshi, \href{https://ieeexplore.ieee.org/abstract/document/9479340}{Model-{Free} {Reinforcement} {Learning} with {Ensemble} for a {Soft} {Continuum} {Robot} {Arm}}, in: 2021 {IEEE} 4th {International} {Conference} on {Soft} {Robotics} ({RoboSoft}), 2021, pp. 141--148.
\newblock \href {http://dx.doi.org/10.1109/RoboSoft51838.2021.9479340} {\path{doi:10.1109/RoboSoft51838.2021.9479340}}.
\newline\urlprefix\url{https://ieeexplore.ieee.org/abstract/document/9479340}

\bibitem{ji2022synthesizing}
Q.~Ji, S.~Fu, K.~Tan, S.~Thorapalli~Muralidharan, K.~Lagrelius, D.~Danelia, G.~Andrikopoulos, X.~V. Wang, L.~Wang, L.~Feng, \href{https://www.sciencedirect.com/science/article/pii/S0736584522000692}{Synthesizing the optimal gait of a quadruped robot with soft actuators using deep reinforcement learning}, Robotics and Computer-Integrated Manufacturing 78 (2022) 102382.
\newblock \href {http://dx.doi.org/10.1016/j.rcim.2022.102382} {\path{doi:10.1016/j.rcim.2022.102382}}.
\newline\urlprefix\url{https://www.sciencedirect.com/science/article/pii/S0736584522000692}

\bibitem{thuruthel2019ModelBased}
T.~G. Thuruthel, E.~Falotico, F.~Renda, C.~Laschi, \href{https://ieeexplore.ieee.org/abstract/document/8531756}{Model-{Based} {Reinforcement} {Learning} for {Closed}-{Loop} {Dynamic} {Control} of {Soft} {Robotic} {Manipulators}}, IEEE Transactions on Robotics 35~(1) (2019) 124--134.
\newblock \href {http://dx.doi.org/10.1109/TRO.2018.2878318} {\path{doi:10.1109/TRO.2018.2878318}}.
\newline\urlprefix\url{https://ieeexplore.ieee.org/abstract/document/8531756}

\bibitem{pei2021improved}
M.~Pei, H.~An, B.~Liu, C.~Wang, An improved dyna-q algorithm for mobile robot path planning in unknown dynamic environment, IEEE Transactions on Systems, Man, and Cybernetics: Systems 52~(7) (2021) 4415--4425.

\bibitem{yu2023safe}
D.~Yu, W.~Zou, Y.~Yang, H.~Ma, S.~E. Li, Y.~Yin, J.~Chen, J.~Duan, Safe model-based reinforcement learning with an uncertainty-aware reachability certificate, IEEE Transactions on Automation Science and Engineering.

\bibitem{bing2023meta}
Z.~Bing, L.~Knak, L.~Cheng, F.~O. Morin, K.~Huang, A.~Knoll, Meta-reinforcement learning in nonstationary and nonparametric environments, IEEE Transactions on Neural Networks and Learning Systems.

\bibitem{ballou2023variational}
A.~Ballou, X.~Alameda-Pineda, C.~Reinke, Variational meta reinforcement learning for social robotics, Applied Intelligence 53~(22) (2023) 27249--27268.

\bibitem{10006753}
L.~Zhu, P.~Peng, Z.~Lu, Y.~Tian, Metavim: Meta variationally intrinsic motivated reinforcement learning for decentralized traffic signal control, IEEE Transactions on Knowledge and Data Engineering 35~(11) (2023) 11570--11584.

\bibitem{li2020random}
J.~Li, X.~Shi, J.~Li, X.~Zhang, J.~Wang, Random curiosity-driven exploration in deep reinforcement learning, Neurocomputing 418 (2020) 139--147.

\bibitem{jiang2024importance}
Y.~Jiang, J.~Z. Kolter, R.~Raileanu, On the importance of exploration for generalization in reinforcement learning, Advances in Neural Information Processing Systems 36.

\bibitem{ibarz2021How}
J.~Ibarz, J.~Tan, C.~Finn, M.~Kalakrishnan, P.~Pastor, S.~Levine, \href{https://doi.org/10.1177/0278364920987859}{How to train your robot with deep reinforcement learning: lessons we have learned}, The International Journal of Robotics Research 40~(4-5) (2021) 698--721.
\newblock \href {http://dx.doi.org/10.1177/0278364920987859} {\path{doi:10.1177/0278364920987859}}.
\newline\urlprefix\url{https://doi.org/10.1177/0278364920987859}

\bibitem{xie2023data}
J.~Xie, H.~Dong, X.~Zhao, \href{https://www.sciencedirect.com/science/article/pii/S0960148123007905}{Data-driven torque and pitch control of wind turbines via reinforcement learning}, Renewable Energy 215 (2023) 118893.
\newblock \href {http://dx.doi.org/https://doi.org/10.1016/j.renene.2023.06.014} {\path{doi:https://doi.org/10.1016/j.renene.2023.06.014}}.
\newline\urlprefix\url{https://www.sciencedirect.com/science/article/pii/S0960148123007905}

\bibitem{cao2021learning}
M.~Cao, R.~Wang, N.~Chen, J.~Wang, A learning-based vehicle trajectory-tracking approach for autonomous vehicles with lidar failure under various lighting conditions, IEEE/ASME transactions on mechatronics 27~(2) (2021) 1011--1022.
\newblock \href {http://dx.doi.org/10.1109/TMECH.2021.3077388} {\path{doi:10.1109/TMECH.2021.3077388}}.

\bibitem{claessens2016convolutional}
B.~J. Claessens, P.~Vrancx, F.~Ruelens, Convolutional neural networks for automatic state-time feature extraction in reinforcement learning applied to residential load control, IEEE Transactions on Smart Grid 9~(4) (2016) 3259--3269.
\newblock \href {http://dx.doi.org/10.1109/TSG.2016.2629450} {\path{doi:10.1109/TSG.2016.2629450}}.

\bibitem{lipton2015critical}
Z.~C. Lipton, J.~Berkowitz, C.~Elkan, \href{http://arxiv.org/abs/1506.00019}{A critical review of recurrent neural networks for sequence learning}, arXiv preprint arXiv:1506.00019 abs/1506.00019.
\newblock \href {http://arxiv.org/abs/1506.00019} {\path{arXiv:1506.00019}}, \href {http://dx.doi.org/arXiv:1506.00019} {\path{doi:arXiv:1506.00019}}.
\newline\urlprefix\url{http://arxiv.org/abs/1506.00019}

\bibitem{sherstinsky2020fundamentals}
A.~Sherstinsky, Fundamentals of recurrent neural network (rnn) and long short-term memory (lstm) network, Physica D: Nonlinear Phenomena 404 (2020) 132306.
\newblock \href {http://dx.doi.org/10.1016/j.physd.2019.132306} {\path{doi:10.1016/j.physd.2019.132306}}.

\bibitem{huang2021continual}
Y.~Huang, K.~Xie, H.~Bharadhwaj, F.~Shkurti, \href{https://ieeexplore.ieee.org/abstract/document/9560793}{Continual {Model}-{Based} {Reinforcement} {Learning} with {Hypernetworks}}, in: 2021 {IEEE} {International} {Conference} on {Robotics} and {Automation} ({ICRA}), 2021, pp. 799--805.
\newblock \href {http://dx.doi.org/10.1109/ICRA48506.2021.9560793} {\path{doi:10.1109/ICRA48506.2021.9560793}}.
\newline\urlprefix\url{https://ieeexplore.ieee.org/abstract/document/9560793}

\bibitem{haarnoja2018soft}
T.~Haarnoja, A.~Zhou, P.~Abbeel, S.~Levine, \href{https://proceedings.mlr.press/v80/haarnoja18b.html}{Soft actor-critic: Off-policy maximum entropy deep reinforcement learning with a stochastic actor}, in: J.~Dy, A.~Krause (Eds.), Proceedings of the 35th International Conference on Machine Learning, Vol.~80 of Proceedings of Machine Learning Research, PMLR, 2018, pp. 1861--1870.
\newline\urlprefix\url{https://proceedings.mlr.press/v80/haarnoja18b.html}

\bibitem{haarnoja2019Soft}
H.~Tuomas, Z.~Aurick, H.~Kristian, T.~George, H.~Sehoon, T.~Jie, K.~Vikash, Z.~Henry, G.~Abhishek, A.~Pieter, L.~Sergey, \href{http://arxiv.org/abs/1812.05905}{Soft actor-critic algorithms and applications}, CoRR abs/1812.05905.
\newblock \href {http://arxiv.org/abs/1812.05905} {\path{arXiv:1812.05905}}, \href {http://dx.doi.org/arXiv:1812.05905} {\path{doi:arXiv:1812.05905}}.
\newline\urlprefix\url{http://arxiv.org/abs/1812.05905}

\bibitem{haarnoja2018composable}
T.~Haarnoja, V.~Pong, A.~Zhou, M.~Dalal, P.~Abbeel, S.~Levine, \href{https://ieeexplore.ieee.org/abstract/document/8460756}{Composable {Deep} {Reinforcement} {Learning} for {Robotic} {Manipulation}}, in: 2018 {IEEE} {International} {Conference} on {Robotics} and {Automation} ({ICRA}), 2018, pp. 6244--6251.
\newblock \href {http://dx.doi.org/10.1109/ICRA.2018.8460756} {\path{doi:10.1109/ICRA.2018.8460756}}.
\newline\urlprefix\url{https://ieeexplore.ieee.org/abstract/document/8460756}

\bibitem{xuezhi2023optimal}
N.~Xuezhi, \href{https://urn.kb.se/resolve?urn=urn:nbn:se:kth:diva-339056}{Optimal {Gait} {Control} of {Soft} {Quadruped} {Robot} by {Model}-based {Reinforcement} {Learning}}, Master's thesis, Dept. Engineering Design, KTH Royal Institute of Technology, Stockholm, Sweden (2023).
\newline\urlprefix\url{https://urn.kb.se/resolve?urn=urn:nbn:se:kth:diva-339056}

\bibitem{biswal2021Development}
P.~Biswal, P.~K. Mohanty, \href{https://www.sciencedirect.com/science/article/pii/S2090447920302501}{Development of quadruped walking robots: {A} review}, Ain Shams Engineering Journal 12~(2) (2021) 2017--2031.
\newblock \href {http://dx.doi.org/10.1016/j.asej.2020.11.005} {\path{doi:10.1016/j.asej.2020.11.005}}.
\newline\urlprefix\url{https://www.sciencedirect.com/science/article/pii/S2090447920302501}

\bibitem{vukobratovic1972Stability}
M.~Vukobratović, J.~Stepanenko, \href{https://www.sciencedirect.com/science/article/pii/0025556472900612}{On the stability of anthropomorphic systems}, Mathematical Biosciences 15~(1) (1972) 1--37.
\newblock \href {http://dx.doi.org/10.1016/0025-5564(72)90061-2} {\path{doi:10.1016/0025-5564(72)90061-2}}.
\newline\urlprefix\url{https://www.sciencedirect.com/science/article/pii/0025556472900612}

\bibitem{hyun2014Higha}
D.~J. Hyun, S.~Seok, J.~Lee, S.~Kim, \href{https://doi.org/10.1177/0278364914532150}{High speed trot-running: {Implementation} of a hierarchical controller using proprioceptive impedance control on the {MIT} {Cheetah}}, The International Journal of Robotics Research 33~(11) (2014) 1417--1445.
\newblock \href {http://dx.doi.org/10.1177/0278364914532150} {\path{doi:10.1177/0278364914532150}}.
\newline\urlprefix\url{https://doi.org/10.1177/0278364914532150}

\bibitem{bertsekas2019reinforcement}
D.~Bertsekas, Reinforcement {Learning} and {Optimal} {Control}, Athena Scientific, Belmont, Massachusetts, 2019.

\bibitem{shao2022Learning}
Y.~Shao, Y.~Jin, X.~Liu, W.~He, H.~Wang, W.~Yang, Learning {Free} {Gait} {Transition} for {Quadruped} {Robots} {Via} {Phase}-{Guided} {Controller}, IEEE Robotics and Automation Letters 7~(2) (2022) 1230--1237.
\newblock \href {http://dx.doi.org/10.1109/LRA.2021.3136645} {\path{doi:10.1109/LRA.2021.3136645}}.

\end{thebibliography}

\end{document}